\documentclass[10pt, a4paper]{article}
\usepackage{lrec2022} 
\usepackage{multibib}
\newcites{languageresource}{Language Resources}
\usepackage{graphicx}
\usepackage{tabularx}
\usepackage{soul}

\usepackage{titlesec}
\titleformat{\section}{\normalfont\large\bfseries\center}{\thesection.}{1em}{}
\titleformat{\subsection}{\normalfont\SmallTitleFont\bfseries\raggedright}{\thesubsection.}{1em}{}
\titleformat{\subsubsection}{\normalfont\normalsize\bfseries\raggedright}{\thesubsubsection.}{1em}{}
\renewcommand\thesection{\arabic{section}}
\renewcommand\thesubsection{\thesection.\arabic{subsection}}
\renewcommand\thesubsubsection{\thesubsection.\arabic{subsubsection}}

\usepackage{epstopdf}
\usepackage[utf8]{inputenc}

\usepackage{xstring}

\usepackage{color}
\usepackage[hyphens]{url}
\usepackage{hyperref}
\usepackage[hyphenbreaks]{breakurl}

\usepackage{booktabs}
\usepackage{enumitem}

\title{Borrowing or Codeswitching? Annotating for Finer-Grained Distinctions in Language Mixing}

\name{Elena Álvarez Mellado\textsuperscript{1}, Constantine Lignos\textsuperscript{2}} 

\address{
    \textsuperscript{1}NLP \& IR Group, School of Computer Science, UNED\\
    \textsuperscript{2}Michtom School of Computer Science, Brandeis University \\
    elena.alvarez@lsi.uned.es, lignos@brandeis.edu \\
}

\abstract{
We present a new corpus of Twitter data annotated for codeswitching and borrowing between Spanish and English.
The corpus contains 9,500 tweets annotated at the token level with codeswitches, borrowings, and named entities.
This corpus differs from prior corpora of codeswitching in that we attempt to clearly define and annotate the boundary between codeswitching and borrowing and do not treat common ``internet-speak'' (\emph{lol}, etc.) as codeswitching when used in an otherwise monolingual context. 
The result is a corpus that enables the study and modeling of Spanish-English borrowing and codeswitching on Twitter in one dataset.
We present baseline scores for modeling the labels of this corpus using Transformer-based language models.
The annotation itself is released with a CC BY 4.0 license, while the text it applies to is distributed in compliance with the Twitter terms of service.
\\
\\
}

\begin{document}

\maketitleabstract

\section{Challenges in Annotating Codeswitching and Borrowing}
\label{sec:shortcomings}

Over the last decade, several initiatives have drawn attention to codeswitching as a challenging NLP task, such as various shared tasks \cite{aguilar-etal-2018-named,molina-etal-2016-overview,solorio-etal-2014-overview} and a recent benchmark \cite{aguilar-etal-2020-lince}.

Prior work on creating datasets of codeswitching has framed the task of language identification for codeswitched utterances as token-level annotation, where every token receives a language identification tag \cite[among others]{maharjan-etal-2015-developing}.
For example, in a collection of English-Spanish codeswitched tweets, tokens in Spanish will be labeled with a language identification tag for Spanish, and tokens in English will be labeled with a language identification tag for English. 

Additional labels have been proposed for use in codeswitching datasets, such as \emph{ambiguous} for words whose language is difficult to determine even in context, \emph{other} for tokens in languages other than the main languages under study, \emph{mixed} for intralexical codeswitching (words that combine morphemes from different languages), \emph{NE} for named entities, and \emph{none} for punctuation marks, emoji, Twitter mentions, etc. 

This repertoire of labels has become the usual tagset in codeswitching shared tasks \cite{aguilar-etal-2018-named,molina-etal-2016-overview,solorio-etal-2014-overview}.
However, this tagset does not take into account that not all other-language inclusions are necessarily a codeswitch. 
In the prior work we examined, we were not able to identify explicit, published definitions or guidelines on what should constitute a codeswitch---or perhaps more crucially, what is \emph{not} a codeswitch.
After all, language mixing can happen in cases that are not necessarily codeswitching.  
Given an utterance that is mostly monolingual, should we assume that any token from another language is a codeswitch, or could it be something else?

Lexical borrowing---the incorporation of single lexical units from one language into another language \cite{haugen1950analysis}---is a process that resembles codeswitching in that it conveys the inclusion of tokens from another language into otherwise monolingual contexts, but it is different in nature \cite{poplack2012myths}.
Codeswitches are usually produced by multilingual individuals, and thus they comply with the grammatical structure of both languages \emph{at the same time}. 
By definition, codeswitches are not integrated into the recipient language, unlike established borrowings. 

On the other hand, lexical borrowings can be produced by monolingual speakers, and they are often accompanied by morphological and phonological modification.
Lexical borrowings comply with the grammatical patterns of the recipient language and can eventually become assimilated into it until the perception of being an inclusion of ``foreign'' origin disappears \cite{lipski2005code}.

With this in mind, it seems unlikely that in a sentence such as \textit{Intentando comprar \underline{online} uno de los nuevos discos duros que sacó Samsung, pero qué lata tener que rellenar tanto formulario},\footnote{``Trying to buy one of the new hard disks released by Samsung online, but what a pain it is to have to fill in so many forms.''} the word \textit{online} could be considered a codeswitch.
Our analysis of existing codeswitching corpora suggests that tokens like \emph{online} have usually been annotated as codeswitches, although in the absence of formal annotation guidelines, it is unclear whether that was intentional or not. On the other hand, a sentence such as \textit{I got it, but prefiero usar mi Dell para cosas sencillas} is clearly a codeswitched utterance.\footnote{``I got it, but I'd rather use my Dell for simple things.''}

Given that lexical borrowing and codeswitching are two distinct linguistic phenomena and the prevalence of lexical borrowings in social media messages \cite{lujan2017analysis,sanou2018anglicismos,stewart-etal-2021-tuiteamos}, we believe that specific guidelines should be provided on how to deal with the codeswitching-borrowing distinction when annotating a dataset for codeswitching.
If codeswitching is the phenomenon of interest, then having a collection of tweets that are rich in other-language inclusions is not sufficient.
As \newcite{poplack2012myths} state, distinguishing codeswitching and borrowing is ``the thorniest issue in the field of contact linguistics today.''

However, previous work on codeswitching dataset design has rarely attempted to establish an explicit difference between intrasentential codeswitching---codeswitching that happens inside one sentence---and lexical borrowing.
Previous work in Hindi by \newcite{bali-etal-2014-borrowing} proposed a statistical approach based on frequency and linguistic typology to capture the distinction between borrowing and mixing, while  \newcite{patro-etal-2017-english} proposed different metrics to compute the likelihood of a word being borrowed being codeswitched.
While many codeswitching datasets include Spanish data,
the codeswitching/borrowing distinction has not been explicitly explored, and to the best of our knowledge, no annotated dataset has made this distinction.

In fact, a glance at the most frequent codeswitches in an English-Spanish codeswitched Twitter dataset \cite{maharjan-etal-2015-developing} reveals that social media abbreviations and well-established internet terms (such as \emph{lol}) are among the most frequent codeswitches annotated.
It is debatable whether these words are in fact true codeswitches, as they can be considered examples of the internet dialect that monolinguals also use rather than evidence of bilinguals switching between languages, as \newcite{maharjan-etal-2015-developing} point out:
\begin{quote}
In the case of abbreviations, some of them such as ``lol'' and ``lmao'' have become social media lingo rather than abbreviations of English words and thus cross language barriers.
\end{quote}

\section{Dataset Description}
\label{sec:data}

With the previously mentioned challenges in mind, we developed a dataset consisting of tweets annotated to reflect codeswitching between Spanish and English as well as borrowing of English into Spanish.
This section describes our annotation guidelines and the process of creating this dataset.
In Section~\ref{subsec:guidelines}, we propose a set of guidelines to distinguish between codeswitching and borrowing that can be followed when annotating a corpus of English-Spanish codeswitched tweets.
In Section~\ref{subsec:selection}, we describe how we reannotated an existing codeswitching dataset to match our guidelines.
Section~\ref{subsec:counts} presents the counts of the resulting dataset and the most frequent tokens per label in the annotation.
The stand-off annotation is released under a Creative Commons CC BY 4.0 license\footnote{\url{https://github.com/lirondos/borrowing-or-codeswitching}}.

\subsection{Annotation guidelines}
\label{subsec:guidelines}

Our annotation uses the following labels:
\begin{itemize}
\item SPA: for tokens in Spanish
\item ENG: for tokens in English
\item OTH: for tokens in languages other than ENG or SPA
\item BOR: for recent borrowings (in English or other languages)
\item ENT: for named entities 
\item N: for punctuation marks, Twitter symbols (such as hashtags and mentions), URLs, etc.
\end{itemize}

In general, tokens in Spanish were annotated as SPA, and tokens in English were annotated as ENG. 
Internet acronyms (such as \textit{lol}) were considered part of the internet jargon and annotated as Spanish whenever they occurred in monolingual Spanish contexts.
Non-Spanish words that appeared in contexts where no other codeswitching was occurring and where there was a strong indication that the word was being borrowed (rather than codeswitched) were annotated as BOR.

Our goal with the borrowing tag was to address relatively recent borrowings into the Spanish language that might be spoken aloud in a normal conversational context, unlike the internet jargon referenced above.

The following were identified as borrowings, differentiating them from short codeswitches embedded in an otherwise monolingual segment:
\begin{enumerate}
    \item English words related to Twitter terminology: such as \emph{Follow Friday}, \emph{tweet}, \emph{follower}, etc.
    \item Technology words: \emph{server}, \emph{hosting}, \emph{user}, \emph{post}, \emph{blog}, \emph{internet}, \emph{template}, \emph{app}, \emph{online}, \emph{chat}.
    \item English words that are already registered in \textit{Diccionario de la lengua española}\footnote{\url{https://dle.rae.es/}} by Real Academia Española \cite{dle}, the general dictionary of standard Spanish compiled by the Royal Spanish Academy, the main prescriptive institution on Spanish language.
    \item English words that are already registered in \textit{Diccionario de Americanismos}\footnote{\url{https://www.asale.org/recursos/diccionarios/damer}} by Asociación de Academias de la Lengua Española \cite{damer}, a specialized dictionary that covers the vocabulary spoken in American Spanish and that has a rich representation of well-established lexical borrowings from English used in Latin America.
    \item English words that are the headword of an entry in Spanish Wikipedia, such as music styles, genres and other cultural things (\emph{hip hop}, \emph{whisky}).
    \item Words that have English origin but were used following Spanish grammatical structure, such as noun-adjective word order (\emph{mensajes offline}, \emph{rating online}).
\end{enumerate}

The intuition behind guidelines \#3--\#5 was that if an English word either appears in some of the main prescriptive dictionaries of Spanish (which tend to be conservative in their approach to language change) or as the headword of a Wikipedia entry, then that would demonstrate that that English word is already well established among monolinguals in at least part of the Spanish-speaking world. 

The motivation for guideline \#6 is that codeswitches by definition normally comply with the grammatical restrictions of both languages involved.
On the other hand, when an item is borrowed from another language---as opposed to codeswitching into the other language---the borrowing will be produced with all the morphosyntactic requisites required by the recipient-language \cite{poplack2012myths}.
Consequently, when we encountered an English word that was used in an otherwise Spanish context and followed all the expectations of Spanish syntax, we annotated it as a borrowing, not a codeswitch.

Our guidelines build on prior work developed to assist annotators when identifying English borrowings in Spanish monolingual texts \cite{alvarez-mellado-2020-annotated,alvarez2020lazaro,alvarez2022detecting}.
That prior work on borrowing annotation was designed exclusively to identify borrowings in Spanish from Spain and used sources whose Spain-centric criteria has been previously pointed out \cite{blanch1995americanismo,fernandez2014lexicografia}, which made them insufficient to identify well-established borrowings in a more geographically diverse collection of texts as our dataset.
Consequently, our guidelines were adapted so that they could account for a more diverse representation of Spanish dialectal varieties than previous guidelines by considering additional lexicographic sources, such as Wikipedia or \textit{Diccionario de Americanismos}.
Our guidelines ensured that English words that are well-established borrowings in many parts of the Spanish speaking Americas (such as \textit{man},\footnote{\url{https://www.rae.es/damer/man}} \textit{nice}\footnote{\url{https://www.rae.es/damer/nice}} or \textit{party}\footnote{\url{https://www.rae.es/damer/party}}) would not be mistaken for codeswitches.  

We have tried to create guidelines for distinguishing borrowing and codeswitching that can be applied somewhat objectively and in the most reliable fashion possible.
As we have stated previously, the distinction is challenging; no annotation guidelines will be perfect, and there is not one universal set of distinctions that all researchers or speakers of the languages involved will agree on.
However, we believe these guidelines can at the very least produce reliable and replicable annotation, and that is an important step forward in this area.

\subsection{Data Selection}
\label{subsec:selection}

To reduce the effort required to create our corpus and place it within existing work on codeswitching, we decided to select an existing corpus already annotated for codeswitching which we could reannotate to match our guidelines.
We selected the codeswitching-dense corpus created by \newcite[hereafter LM2013]{lignos2013toward} for several reasons.

First, all tweets in the original LM2013 corpus had been annotated by two annotators and the original decisions made by each annotator were readily available, allowing us to identify and revisit any disagreements between the two annotators.

Second, the LM2013 annotators had been instructed to annotate the data to attempt to differentiate borrowing and codeswitching.
While the annotation does not have a borrowing tag, the annotators were instructed not to annotate borrowed words as codeswitches.
In the LM2013 annotation process, this was referred to as the ``monolingual grandparent'' rule: if it is unclear whether a word is a borrowing or codeswitch, if one's monolingual grandparent would recognize it as part of their language, do not annotate it as a codeswitch.
While this was intended to draw upon the life experiences of many bilinguals in that they are able to identify which words are understood by monolingual speakers, because it referred to an older relative it may have unintentionally had the effect of causing the annotators to annotate more recent borrowings as codeswitches.\footnote{Perhaps future work would be better served by a ``monolingual cousin'' rule instead.}

Third, the data used by LM2013 is primarily Spanish,\footnote{The tweets were selected from the Spanish subset of \newcite{burger-etal-2011-discriminating}, which was designed to be a general Spanish Twitter dataset created using automatic language identification without any intent to study language mixing. However, that dataset includes a significant amount of Spanish-English codeswitched tweets, making it a useful source for codeswitching data. The subset of the dataset selected by LM2013 was chosen to contain codeswitched tweets at a higher rate than the rest of the corpus.} which gives more opportunities to annotate borrowings into Spanish, especially those from English which is responsible for a large number of emergent borrowings \cite{furiassi2012anglicization,gorlach2002english}.

The LM2013 dataset was reannotated following the guidelines from section \ref{subsec:guidelines} by a native speaker of Spanish that had a background in linguistics and had previous experience in linguistic annotation. Our reannotation differs from the original annotation of LM2013 in several ways.
We completed adjudication of all 9,500 tweets annotated by LM2013.
Due to limited adjudicator availability, they had only completed adjudication of 7,018 tweets, with the remainder having unresolved disagreements.
The LM2013 dataset assigned language tags to named entities---names were annotated Spanish or English based on the surrounding language, not the source language of the name---but we found that this annotation was unreliable and we removed it, making our data more consistent with other codeswitching datasets that do not make this distinction.

LM2013's original annotation also distinguished between two types of entities: works of art (e.g., book titles) and other named entities.
We removed this distinction as it was inconsistently annotated and most subsequent annotation either uses standard named entity types (PER, LOC, ORG, etc.) or a general ``named entity'' tag \cite{maharjan-etal-2015-developing} like our annotation.

Finally, we explicitly annotated borrowings as such.
The original LM2013 annotation did not do so and the LM2013 annotators were instructed to not treat borrowings as codeswitches, but the annotators followed this advice inconsistently.
While the annotators were not consistent, the fact that they were instructed to make this distinction was useful, as it resulted in inconsistencies that we could easily identify in adjudication.

As discussed in Section~\ref{sec:shortcomings}, a major difference between our annotation and previous codeswitching corpora that we are aware of (including the original LM2013 annotation) is that we do not count ``internet-speak'' as a codeswitch into English, reflecting the fact that monolingual communities may use internet shorthand adapted from English even in purely monolingual non-English contexts.

In order to identify borrowing candidates that could fall under the definition established in Section~\ref{subsec:guidelines}, we manually looked for frequently used English borrowings, checked for inter-annotator disagreements (for example \textit{Me mandas tu \underline{mail} y lo envío},\footnote{``Send me your email address and I will sent it to you.''}) and automatically flagged inconsistencies in how certain tokens were sometimes labeled as SPA and sometimes annotated as ENG (such as \textit{man}, \textit{server} or \textit{tweet}) across the whole corpus, even if annotators agreed on the individual annotations.
This approach proved to be successful and revealed a high number of tokens that were actually borrowings and not true codeswitches (see Section~\ref{subsec:counts}).  

The result of our reannotation process is a corpus that makes the distinction between borrowing and codeswitching and has a narrower definition of codeswitching than previous work, allowing for investigation of a finer-grained linguistic phenomenon.

\subsection{Data counts}
\label{subsec:counts}

\begin{table}[tb]
\centering
\begin{tabular}{lr}
\toprule
Tag & Tokens \\
\midrule
SPA & 134,110 \\
N   &  39,280 \\
ENT &  15,373 \\
ENG &   6,819 \\
BOR &   2,857 \\
OTH &     267 \\
\midrule
Total & 198,706 \\
\bottomrule
\end{tabular}
\caption{Token counts by label}
\label{tab:counts}
\end{table}

The corpus consists of 9,500 tweets that were annotated following the guidelines established in Section~\ref{subsec:guidelines}.
Out of those 9,500 tweets, 2,017 contained at least one borrowing, 2,495 contained at least one codeswitch, and 403 contained both a borrowing and a codeswitch.
In total there are 198,706 tokens in the corpus, and counts for each label are given in Table~\ref{tab:counts}.
The largest sources of tokens labeled OTH belong to tweets with codeswitches into languages that are neither Spanish nor English, typically Portuguese and Catalan.

Table~\ref{tab:freq} summarizes the most frequent tokens per label.
As there are essentially zero purely-English tweets in this dataset, the counts of English tokens reflect the count of tokens of codeswitches into English.
The most frequent tokens show a noticeable difference with the most frequent tokens from some other codeswitched data counts.
While \newcite{maharjan-etal-2015-developing} reported internet terms and abbreviations such as \textit{lol}, \textit{lmao} or \textit{idk} among the most frequent codeswitched tokens, the most frequent English tokens in our dataset are exclusively function words distributed in an approximately Zipfian fashion.
This matches what we would expect in a fully monolingual English corpus and is similar to the distribution that tokens labeled as Spanish follow.

This suggests that what we are actually seeing labeled as English are true English sub-utterances---true codeswitches---and not the well-established borrowings that we find under the most frequent BOR column, or the ``internet-speak'' that we did not annotate as either a borrowing or codeswitch.
However, it should be noted that the comparison of the amount and complexity of codeswitching in the data with other datasets is difficult because a quantitative codeswitching measure like that of \newcite{gamback2016comparing} cannot be meaningfully applied across corpora with different annotation guidelines.

\begin{table*}[tb]
\centering
\begin{tabular}{lr|lr|lr}
\toprule
Spanish & Count & English & Count & Borrowing & Count \\
\midrule
 de & 6,175 &  the & 119     & blog & 231 \\
   que & 4,298     &   I & 105       &    post & 130  \\
  y & 3,281       &   you & 84     &    web & 107 \\
  el & 3,257      &   to & 81      &    internet & 106 \\
  a & 3,178       &   my & 69      &     followers & 55 \\
  la & 3,175      &   a & 69       &     online & 41   \\
   en & 3,120      &  it & 62      &     blogs & 41   \\
   no & 2,410      &  is & 60      &     software & 33  \\       
   me & 2,086      &  in & 56      &     Internet & 33   \\      
   es & 1,764      &  on & 49      &     timeline & 32   \\     
\bottomrule
\end{tabular}
\caption{The ten most frequent tokens in the dataset for each label: Spanish, English, and borrowing}
\label{tab:freq}
\end{table*}

\subsection{Limitations}
\label{sec:limitations}

Our goal for creating this corpus is to study a narrowly-scoped slice of multilingual language usage on Twitter---how Spanish speakers codeswitch into English, how they use borrowings and how these two phenomena intertwine---with the cleanest and most carefully annotated dataset possible.
This corpus is not designed to be an evenly-balanced English-Spanish codeswitching dataset.
It is primarily a Spanish dataset, mainly composed of Spanish tweets that may have an English codeswitch or a borrowing.

While our guidelines seek to establish a division as clearly as possible, the distinction between codeswitching and borrowing is fuzzy and is far from being solved.
Many gray areas exist between social media jargon, nonce borrowing, and true codeswitching.

Another limitation is that as the original data selection process used generic Spanish language identification---\newcite{burger-etal-2011-discriminating} do not provide any details---we do not know the geographic distribution of the speakers in the dataset and what dialects of Spanish and English they used.
While we do believe this is a useful dataset for studying codeswitching and borrowing, caution should be exercised before drawing any global conclusions about the usage and mixing of either language.

\section{Baseline Models}
\label{sec:models}

\begin{table*}[tb]
\centering
\begin{tabular}{lrrrr}
\toprule
Model & Accuracy & Precision & Recall & F1 \\
\midrule
mBERT           & 96.88          & \textbf{96.69} & \textbf{96.61} & \textbf{96.65}   \\
BETO            & \textbf{96.91} & \textbf{96.69} &          96.60 & 96.64   \\
RoBERTa-BNE     & 93.73          &   93.19        &          93.23 & 93.21   \\
RoBERTa Twitter & 93.39          &   92.82        &          92.86 & 92.84  \\
\bottomrule
\end{tabular}
\caption{Accuracy and micro-averaged precision, recall, and F1 score for baseline models (results from a single run)}
\label{tab:models}
\end{table*}

We evaluated how well popular models perform in predicting the annotated labels of our dataset.
The data was divided into a standard training, development, and test set split in 80/10/10 proportions.

We evaluated four Transformer-based models on the dataset: 
\begin{itemize}
    \item mBERT: multilingual BERT, trained on Wikipedia in 104 languages \cite{devlin-etal-2019-bert}
    \item BETO base cased model: a BERT-based model trained on Spanish \cite{CaneteCFP2020}
    \item RoBERTa BNE: a RoBERTa based model trained on Spanish \cite{gutierrezfandino2021spanish}
    \item RoBERTa Twitter: a RoBERTa based model trained on English Twitter data \cite{barbieri-etal-2020-tweeteval}
\end{itemize}

The aim behind this selection of models was to assess how Transformer-based models that had been trained on different types of text would perform on codeswitched data. 

The only multilingual model that we included was mBERT. 
BETO is a BERT-based model trained on a diverse set of international Spanish texts from different origins, such as OpenSubtitles, Global Voices, and the United Nations \cite{canete}. 
RoBERTa BNE is a RoBERTa-based model that has been trained exclusively on data crawled from .es websites---those using the top-level domain for Spain---by the National Library of Spain. 
This means that while BETO training data is smaller (3 billion tokens vs BNE's 135 billion tokens), the model has probably been exposed to a more varied representation of the different varieties of Spanish spoken around the globe, which includes the area of Spanish where codeswitching may be more prevalent. 
Finally, we added a RoBERTa-based model trained on Twitter data under the assumption that an English monolingual model trained on social media text could potentially do better than other models trained on other genres. 

All models were run using the Transformers library by HuggingFace \cite{wolf-etal-2020-transformers} with the same default (untuned) hyperparameters: 3 epochs, batch size 32, and a maximum sequence length of 256.\footnote{\url{https://github.com/huggingface/transformers/tree/master/examples/pytorch/token-classification}} 

Table~\ref{tab:models} shows that both mBERT and BETO were the best performing models and achieved similar scores, with an F1 of 96.6.
While we have bolded the highest number in each column, we have no reason to believe that any differences between the performance of mBERT and BETO are meaningful.
Interestingly, the RoBERTa-based model trained on the crawled data by the National Library of Spain performed worse, despite being a larger model.
This seems to suggest the importance that a diverse set of training material may have when modeling certain linguistic phenomena that are impacted by geographical and dialectal variation, such as codeswitching and borrowing.

While we cannot directly compare our scores to other results as they are the first results on this dataset, these results are similar to those obtained on other datasets.
For example, in the LinCE benchmark baseline \cite{aguilar-etal-2020-lince},\footnote{\url{https://ritual.uh.edu/lince/}} for the language identification task on Spanish-English codeswitching data the F1 scores were 94.16 with a BiLSTM, 98.12 with ELMo, and 98.53 with mBERT.
The mBERT baseline F1 for our dataset is slightly lower at 96.65.
One possible explanation is that our task is slightly more difficult for the model as it must make the borrowing-codeswitching distinction, but further experimentation and tuning would be required to support that claim.

The token-level accuracy of approximately 96.9 for mBERT and BETO baseline matches the accuracy of 96.9 reported by \newcite{lignos2013toward} on their data with a heuristic-based tagging approach.
However, their evaluation excluded any tokens with named entity tags and their data was not annotated for borrowing, so it is likely their approach would get lower accuracy if evaluated on our reannotation.



\section{Conclusion}

We have introduced a new dataset of tweets annotated both with lexical borrowings and Spanish-English codeswitches.
The annotation builds on previous approaches to codeswitching dataset creation, but distinguishes lexical borrowing from true codeswitching.
This distinction has been previously pointed out as crucial in the contact linguistics literature, but has not been made in previous codeswitching datasets.
We have experimented with different Transformer-based models for the task of language identification and compared results in our dataset to previous work on other codeswitching datasets.


\section{Bibliographical References}
\label{reference}

\bibliographystyle{lrec2022-bib}
\bibliography{main}



\end{document}